\documentclass[10pt,twocolumn,letterpaper]{article}

\usepackage{cvpr}      
\usepackage{multirow}
\usepackage{comment}

\usepackage[dvipsnames]{xcolor}

\definecolor{cvprblue}{rgb}{0.21,0.49,0.74}
\usepackage[pagebackref,breaklinks,colorlinks,citecolor=cvprblue]{hyperref}
\def\paperID{15769} 
\def\confName{CVPR}
\def\confYear{2024}

\title{Domain-Rectifying Adapter for Cross-Domain Few-Shot Segmentation}

\author{
Jiapeng Su\textsuperscript{1}\thanks{Both authors contributed equally.}, Qi Fan\textsuperscript{2}\footnotemark[1],
Guangming Lu\textsuperscript{1}, Fanglin Chen\textsuperscript{1}, Wenjie Pei\textsuperscript{1}\thanks{Corresponding author.} \\
\textsuperscript{1}Harbin Institute of Technology, Shenzhen\quad 
\textsuperscript{2}Nanjing University \\
{\tt\small MattSu@163.com, fanqics@gmail.com, luguangm@hit.edu.cn,} \\
{\tt\small chenfanglin@hit.edu.cn, wenjiecoder@outlook.com}
}

\begin{document}

\maketitle

\begin{abstract}

Few-shot semantic segmentation (FSS) has achieved great success on segmenting objects of novel classes, supported by only a few annotated samples. 
However, existing FSS methods often underperform in the presence of domain shifts, especially when encountering new domain styles that are unseen during training.
It is suboptimal to directly adapt or generalize the entire model to new domains in the few-shot scenario.
Instead, our key idea is to adapt a small adapter for rectifying diverse target domain styles to the source domain.
Consequently, the rectified target domain features can fittingly benefit from the well-optimized source domain segmentation model, which is intently trained on sufficient source domain data.
Training domain-rectifying adapter requires sufficiently diverse target domains.
We thus propose a novel local-global style perturbation method to simulate diverse potential target domains by perturbating the feature channel statistics of the individual images and collective statistics of the entire source domain, respectively.
Additionally, we propose a cyclic domain alignment module to facilitate the adapter effectively rectifying domains using a reverse domain rectification supervision.
The adapter is trained to rectify the image features from diverse synthesized target domains to align with the source domain.
During testing on target domains, we start by rectifying the image features and then conduct few-shot segmentation on the domain-rectified features.
Extensive experiments demonstrate the effectiveness of our method, achieving promising results on cross-domain few-shot semantic segmentation tasks. Our code is available at \href{https://github.com/Matt-Su/DR-Adapter}{https://github.com/Matt-Su/DR-Adapter}.


\end{abstract}

\section{Introduction}
\label{sec:intro}
\graphicspath{{fig/}}


Benefiting from well-established large-scale datasets~\cite{benenson2019large,lin2014microsoft}, numerous semantic segmentation methods~\cite{long2015fully,chen2017deeplab,ronneberger2015u} have undergone rapid development in recent years. However, obtaining enough labeled data is still a challenging and resource-intensive process, particularly for tasks like instance and semantic segmentation. Unlike machine learning approaches, human capacity to recognize novel concepts from limited examples fuels considerable research interest. Hence, few-shot segmentation (FSS) is proposed to meet this challenge, developing a network that generalizes to new domains with limited annotated data.

\begin{figure}[t]
  \centering
   \includegraphics[width=1.0\linewidth]{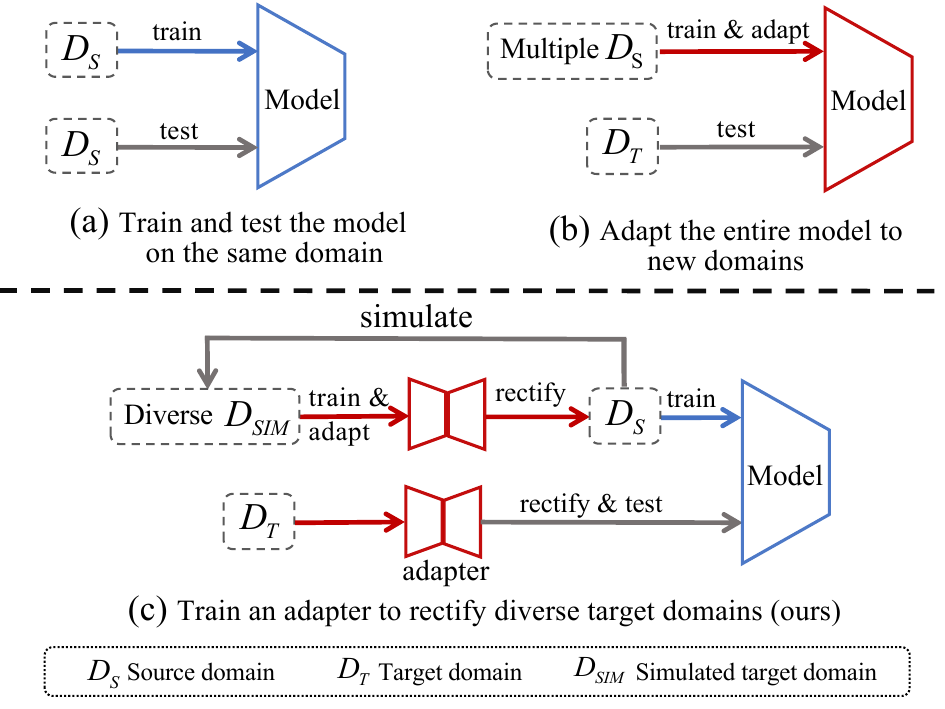}
       \vspace{-0.15in}

   \caption{The comparison of our method with other approaches.
   (a) Traditional few-shot segmentation (FSS) methods train and test the model on the same domain. (b) Most domain generalization (DG) methods leverages multiple source domains to train and adapt the large-parameter model simultaneously.
   (c) In contrast to conventional DG methods, we propose using a lightweight adapter as a substitute. This adapter is designed to adapt to various domain data, thereby decoupling domain adaptation from the source domain training process.}
   \label{fig:comparison}
       \vspace{-0.15in}

\end{figure}

Nonetheless, most existing few-shot segmentation methods~\cite{nguyen2019feature,lu2021simpler,zhang2021self,zhang2019canet,tian2020prior,li2021adaptive,zhang2021few,min2021hypercorrelation,fan2022self} often exhibit subpar performance when confronted with domain shifts~\cite{liu2020feature,saito2021tune,yue2021prototypical}. The cross-domain few-shot segmentation (CD-FSS) is thus proposed for generalizing few-shot segmentation models from the source domain to other domains~\cite{lei2022cross,wang2022remember}.
CD-FSS trains the model solely on the source domain, and generalizes the trained model to segment object of novel classes in a separate target domain, supported by few-shot samples.


Domain adaptation(DA) and domain generalization(DG) are closely related to cross-domain few-shot segmentation. However, DA methods require unlabeled training data from the target domain. DG aims to generalize models trained in the source domain to various unseen domains, often requiring extensive training data from multiple source domains.
Consequently, DA/DG methods typically adapt the entire model to new domains, leveraging substantial domain-specific data.
Similarly, most existing CD-FSS methods adapt the entire model to target domains.
However, in few-shot learning, the scarcity of training data can lead to overfitting when directly adapting the entire model.
Rather than generalizing the entire model, our approach focuses on adapting a compact adapter to rectify diverse target domain features to align with the source domain.
Once rectified to the source domain, target domain features can effectively utilize the well-trained source domain segmentation model, which is intently optimized using extensive source domain data.
Figure~\ref{fig:comparison} shows the difference among our method and conventional FSS and domain generalization methods. 

\begin{figure}[t]
  \centering
   \includegraphics[width=1.1\linewidth]{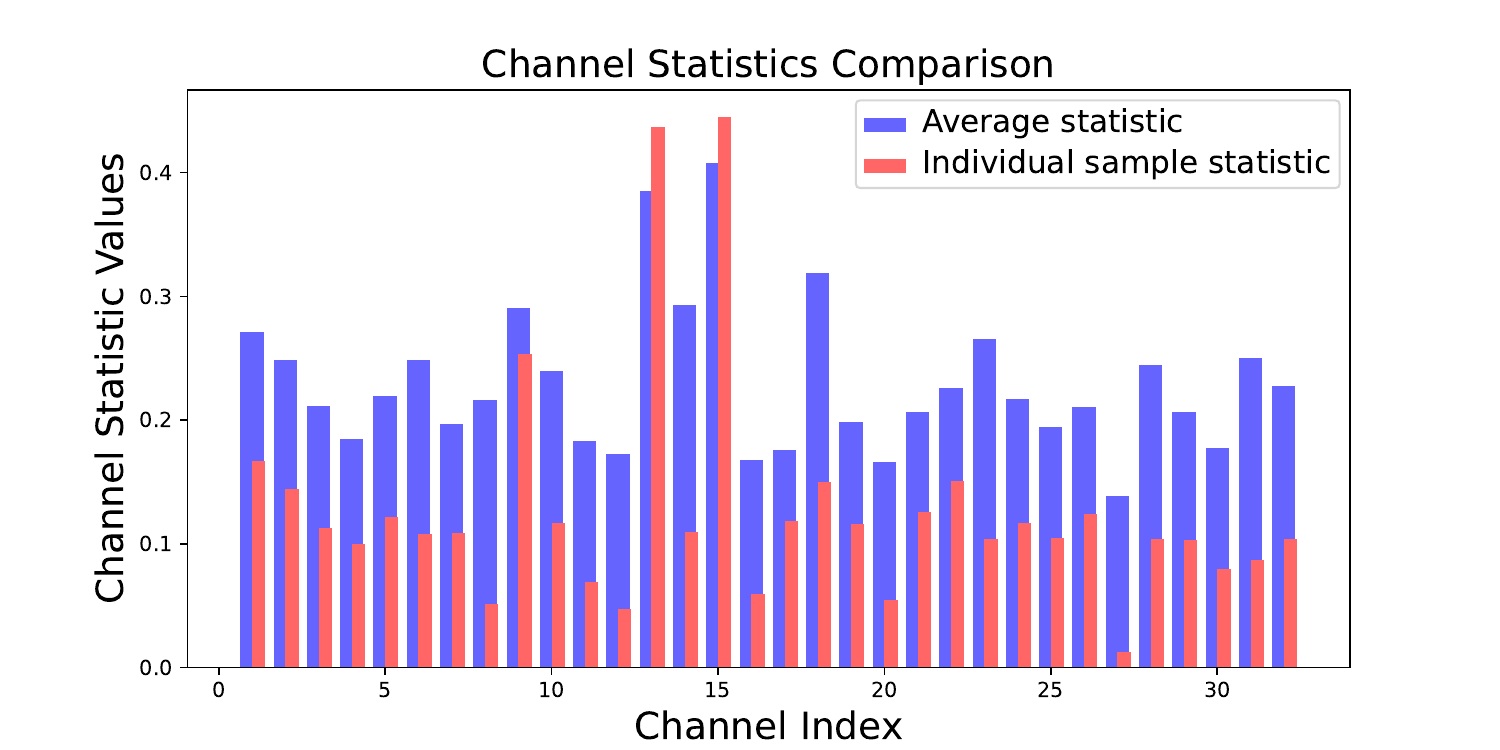}
    \vspace{-0.25in}

   \caption{We show the feature channel statistic of an individual sample's statistic and the average statistic across the dataset on the pretrained backbone at stage 1. The average statistics exhibit a smoother profile compared to that of an individual sample, allowing for the application of more substantial noise to the feature with the smoother statistics.}
   \label{fig:channel_scale}
       \vspace{-0.15in}

\end{figure}

Training a domain-rectifying adapter requires extensive data of diverse target domains.
The straightforward feature-level domain synthesis method can effectively generate diverse potential target domains by randomly perturbing feature channel statistics.
We can diversify the synthesized domain styles by increasing the magnitude of perturbation noises.
However, as shown in Figure \ref {fig:channel_scale}, 
some feature channels in individual images exhibit very low activation values.
These small feature channel statistic values result in the corresponding channels suffering from limited style synthesis.
Merely increasing the perturbation noises may lead to model collapse, where highly activated channels are excessively perturbed.
Consequently, we propose a novel local-global style perturbation method to generate diverse potential target domain styles.
Our local style perturbation module generates new domains by perturbing the feature channel statistics of individual images, similar to DG methods.
Our global style perturbation module effectively diversifies the synthesized styles by leveraging the collective feature statistics of the entire source domain.
Dataset-level feature statistics are estimated through momentum updating on the entire source domain dataset.
Our local and global style perturbation modules collaboratively generate diverse and meaningful domain styles.

The perturbed feature channel statistics represent diverse potential styles, which are then input into the adapter to train the domain-rectifying adapter.
The adapter predicts two rectification vectors to rectify the perturbed feature channel statistics to their original values.
Additionally, we propose a cyclic domain alignment module to assist the adapter in learning to effectively rectify diverse domain styles to align with the source domain.
Once rectified, the feature channel statistics will collaborate with the normalized feature map to train the segmentation model.
During inference, we can directly use the domain-rectifying adapter to align the image features with the source domain and then input them into the well-trained source domain model for segmentation.
In summary, our contributions are:
\begin{itemize}
	\item We introduce a novel domain-rectifying method for cross-domain few-shot segmentation, employing a compact adapter to align diverse target domain features with the source domain, mitigating overfitting in limited training data scenarios.
 
	\item We propose a unique local-global style perturbation module that generates diverse target domain styles by perturbing feature channel statistics at both local and global scales, enhancing model adaptability to various target domains.
 
	\item To enhance domain adaptation, we introduce a cyclic domain alignment loss that helps the domain-rectifying adapter align diverse domain styles with the source domain.
\end{itemize}

\section{Related Work}

\subsection{Few-Shot Segmentation}
Few-shot semantic segmentation~\cite{zhang2020sg,zhang2019canet,wang2019panet,tian2020prior,liu2020part,yang2021mining,liu2022learning,liu2020dynamic,liu2020crnet}, using a limited number of labeled support images, predicts dense masks for query images. Previous methods primarily adopted a metric-based paradigm~\cite{dong2018few}, improved in various ways, and fell into two main categories: prototype-based and matching-based approaches. Motivated by PrototypicalNet~\cite{snell2017prototypical}, prototype-based methods extract prototypes from support images to guide query object segmentation. Most studies concentrate on effectively utilizing limited support images to obtain more representative prototypes. Recent studies~\cite{zhang2019pyramid,zhang2021few} emphasize that a single prototype often fails to represent an entire object adequately. To address this, methods such as ASGNet~\cite{li2021adaptive} and PRMMs~\cite{tian2020prior} explore using multiple prototypes to represent the overall target.

On the other hand, matching-based methods~\cite{lu2021simpler,min2021hypercorrelation,vinyals2016matching} concatenate support and query features, subsequently inputting the concatenated feature map into CNN or transformer networks. This process explores the dense correspondence between query images and support prototypes. Recently, researches~\cite{peng2023hierarchical,shi2022dense} has focused on leveraging  pixel-to-pixel similarity maps for effective support prototype generation and query feature enhancement.

\subsection{Domain Generalization}
Domain Generalization (DG) targets at generalizing models to diverse target domains, particularly when target domain data is inaccessible during training. Existing domain generalization methods fall into two categories: learning domain-invariant feature representations from multiple source domains~\cite{du2020learning,ghifary2015domain,motiian2017unified,wang2019learning} and generating diverse samples via data or feature augmentation~\cite{carlucci2019domain,shankar2018generalizing,volpi2018generalizing,zhou2020learning}. The core idea of learning domain-invariant features is to leverage various source domains to learn a robust feature representation. Data or feature augmentation aims to increase the diversity of training samples to simulate diverse new domains.

Domain generalization is particularly challenging in few-shot settings, as the target domain substantially differs from the source domains in both domain style and class content. Unlike popular DG methods generalizing the entire model, we train a small adapter to rectify the target domain data into the source domain style for model generalization.

\subsection{Cross-domain Few-Shot segmentation}

Recently, cross-domain few-shot segmentation has received increasing attention. PATNet~\cite{lei2022cross} proposes a feature transformation layer to map query and support features from any domain into a domain-agnostic feature space. RestNet~\cite{huang2023restnet} addresses the intra-domain knowledge preservation problem in CD-FSS. RD~\cite{wang2022remember} employs a memory bank to restore the meta-knowledge of the source domain to augment the target domain data.
Unlike previous CD-FSS methhods, our method directly learns two rectification parameters for effective domain adaptation, eliminating the needs of restoring source domain styles.


\section{Methodology}

\noindent{\bf Problem Setting} Cross-Domain Few-Shot Segmentation (CD-FSS) aims to apply the source domain trained few-shot segmentation models to diverse target domains. 
The CD-FSS model is typically trained using episode-based meta-learning paradigm ~\cite{vinyals2016matching,fan2020few}.
The training and testing data both consist of thousands of randomly sampled episodes, including $K$ support samples and one query image. 
The model first extracts the support prototype and query feature from each training episode, and then performs pixel-wise feature matching between the support prototype and query feature to predict the query mask.
The support prototype is typically a feature vector aggregating the object features of all support images.
Once trained, the model is directly applied to various target domains.

\begin{figure*}[]
	\centering
	\includegraphics[width=1\linewidth]{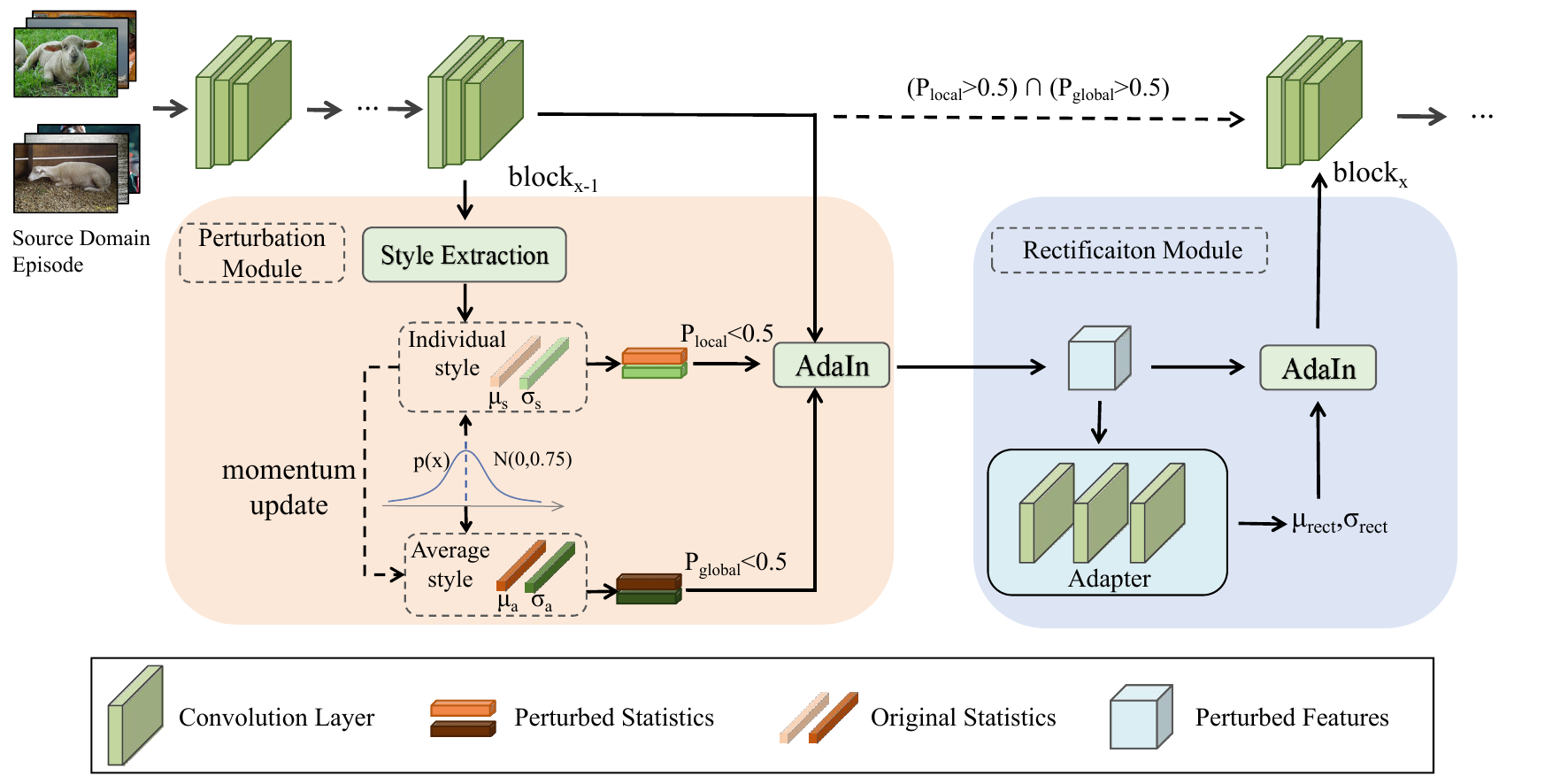}
     \vspace{-0.2in}
	\caption{Overview of our cross-domain few-shot segmentation approach. Our method consists of two modules: a feature perturbation module and a feature rectification module. The former is used to generate simulated domain features, while the latter trains the adapter by restoring the features to their original states. During the perturbation process, we employ both local and global perturbations, controlled by two different probabilities $P$ to decide if a feature is perturbed. Note that when both probabilities exceed 0.5, the entire backbone undergoes standard training. During testing, we treat target domain features as perturbed features and directly rectify them using the adapter.}
	\label{fig:framwork}
    \vspace{-0.1in}
\end{figure*}

\noindent{\bf Method Overview} Our key idea is to train a adapter to rectify diverse target domain styles to the source domain, and leverage the well-trained source domain segmentation model to process the rectified target domain features for accurate few-shot segmentation.
The crux is to align diverse potential target domain distributions to the source domain distribution.
To train the domain-rectifying adapter, we thus synthesize various target domain styles by perturbing the feature channel statistics of the source domain training images.
And the adapter is trained to rectify the synthesized feature styles to the source domain style.
During inference, the adapter can be directly applied on the target domain features to rectify their domain styles, and the subsequent segmentation model can process the rectified support and query features for few-shot segmentation.
he overall framework of our approach is illustrated in Figure \ref{fig:framwork}.




\subsection{Local Domain Perturbation}

Previous works~\cite{fan2022towards,zhou2021domain} show that perturbing feature channel statistics can effectively synthesize diverse domain styles and meanwhile preserves the image contents.
We thus synthesize various domain styles by injecting gaussian noises into feature channel statistics of source domain images.

Given a feature map $F_{o} \in \mathcal{R}^{B \times C \times H \times W}$, 
we first compute the feature channel statistics, \ie, mean $\mu _{o}$ and variance $\sigma_{o}$ along each channel dimension:
\begin{align}
\mu_{o}(F_o) &= \frac{1}{HW} \sum_{h=1}^{H} \sum_{w=1}^{W} F_o, \\
\sigma_{o}(F_o) &= \sqrt{\frac{1}{HW} \sum_{h=1}^{H} \sum_{w=1}^{W} \left(F_o - \mu_{o}(F_o)\right)^{2} + \epsilon},
\end{align}
where $\mu_{o}, \sigma_{o} \in \mathcal{R}^{B \times C}$, $\epsilon$ is a small constant for numerical stability, $B$, $C$, $H$, and $W$ represent the batch size, channel dimension, height, and width of the feature map.

Then, we leverage two perturbation factors $\alpha$ and $\beta$ to control the gaussian noise injection process for $\mu_{o}$ and $\sigma_{o}$. The noise vectors, sharing the same dimension as $\mu_{o}$ and $\sigma_{o}$, are used to compute the perturbed mean $\mu_{p}$ and variance $\sigma_{p}$:
\begin{equation}
\begin{aligned}
\mu_{p} &= (1 + \alpha) \mu_{o}, \\
\sigma_{p} &= (1 + \beta) \sigma_{o}.
\end{aligned}
\end{equation}

We can obtain the perturbed feature map $F_p$ by replacing the feature channel statistics $\{
\mu_o, \sigma_o\}$ of the original feature map $F_o$ with the perturbed channel statistics $\{
\mu_p, \sigma_p\}$ using the Adaptive Instance Normalization formula ~\cite{huang2017arbitrary}:
\begin{equation}
F_{p} = \sigma_{p} \frac{F_{o} - \mu_{o}}{\sigma_{o}} + \mu_{p}.
\end{equation}

Within each episode, the support and query features share the same perturbation factors. The above equations can be further simplified to the following expression:
\begin{equation}
\label{eq_fp}
F_{p} = (1 + \beta) F_{o} + (\alpha - \beta) \mu_{o}.
\end{equation}

We call this feature channel statistic perturbation method as local domain perturbation, as it is enabled on individual images with probability $P_{\text{local}}$.


\subsection{Global Domain Perturbation}

We need to bound the local domain perturbation to prevent potential training collapse caused by the aggressive perturbation noises.
However, insufficient domain perturbation may lead the domain-rectifying adapter to underperform when encountering new domain styles.
The local domain perturbation method is trapped in the stability and performance dilemma.
We thus propose a novel global domain perturbation by leveraging the global style statistics of the entire dataset to facilitate the domain style synthesis.
The dataset's global style statistics exhibit better perturbation stability when leveraging aggressive perturbation noises to synthesize meaningful target domain styles for sufficient style diversity.

We first compute the feature channel statistics $\mu_o$ for individual images and then progressively update the global style statistics through momentum updating:
\begin{equation}
\mu_{\text{datum}} = \lambda \mu_{\text{datum}} + (1-\lambda)\mu_{o},
\end{equation}
where $\lambda$ is the momentum updating factor.
Then we can perform the global domain perturbation by replacing the image feature channel statistics $\mu_o$ in equation~\ref{eq_fp} with the global style statistics $\mu_{\text{datum}}$.
This global domain perturbation is randomly enabled with probability $P_{\text{global}}$.

\subsection{Domain Rectification Module}

The domain rectification module leverages an domain-rectifying adapter to rectify the target domain feature channel statistics to the source domain. The adapter takes as input the perturbed features and predicts two rectification vectors $\{\alpha_{rect}, \beta_{rect}\}$ to rectify the feature channel statistics of the perturbed feature map $F_p$ as the rectified feature channel statistics $\{\mu_{rect}, \sigma_{rect}\}$. 
\begin{equation}
\begin{aligned}
    \mu_{rect} &= (1+\alpha_{rect})\mu_{p}, \\
    \sigma_{rect} &= (1+\beta_{rect})\sigma_{p}.
\end{aligned}
\end{equation}

Then we leverage the AdaIN function to generate the rectified feature map $F_{\text{rect}}$ based on the perturbed feature map $F_{p}$ and the rectified feature channel statistics $\{\mu_{rect}, \sigma_{rect}\}$:
\begin{align}
    \mathrm{F_{rect}} = \left(1+\beta_{rect}\right)\sigma_{p} \frac{\mathrm{F_{p}-\mu_{p}}}{\sigma_{p}} + \left(1+\alpha_{rect}\right)\mu_{p},
\end{align}
which can be further simplified as:
\begin{align}
    \mathrm{F_{rect}} = \left(1+\beta_{rect}\right)\mathrm{F_{p}}+\left ( \alpha _{rect}-\beta _{rect}   \right )\mu _{p}.
\end{align}

We expect the adapter can adaptively predict the rectification factors $\{\alpha_{rect}, \beta_{rect}\}$ to rectify the perturbed features corresponding to diverse potential target domains.
Consequently, during inference, we can leverage the adapter to rectify the target domain features to the source domain, and the rectified features can fittingly benefit from the well-trained source domain model for satisfactory few-shot segmentation results.

\begin{figure}[t]
  \centering
   \includegraphics[width=0.9\linewidth]{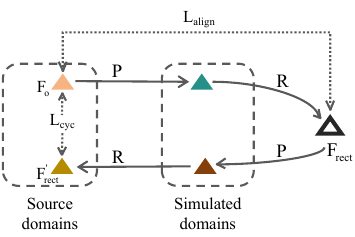}
       \vspace{-0.1in}

   \caption{The process of cycle alignment, where 'P' denotes perturbation and 'R' stands for rectification.}
   \label{fig:cycle}
       \vspace{-0.15in}

\end{figure}

\subsection{Cyclic Domain Alignment}
Our goal is enabling the adapter to rectify the perturbed features back to the source domain space. Insufficient supervision during this process may lead the adapter to rectify the features into an unknown space. Therefore, in addition to utilizing the standard Binary Cross-Entropy (BCE) loss for supervision, we propose incorporation of a cyclic alignment loss to constrain the adapter.

After obtaining the rectified feature $F_{rect}$, we further perturb the $F_{rect}$ with the same noise $\alpha$ and $\beta$ to get a new perturbed feature $F_{rect}^{p}$. This perturbed image $F_{rect}^{p}$ is then input into the adapter for a reverse rectification, resulting in $F_{rect}^{'}$. If the adapter can map features back to the source domain space, the style of  $F_{rect}^{'}$ should closely match that of $F_{o}$. The cycle process is shown in figure \ref{fig:cycle}. Consequently, we align the statistics between original feature and the cyclically rectified feature:
\begin{equation}
\begin{aligned}
    L_{\text{cyc}} &= \frac{1}{C} \sum_{c} \left( \left| \mu\left( F_{o} \right) - \mu\left( F'_{\text{rect}} \right) \right| \right.\\
    &\quad + \left. \left| \sigma\left( F_{o} \right) - \sigma\left( F'_{\text{rect}} \right) \right| \right).
\end{aligned}
\end{equation}

We add constraint to the statistics between  $F_{o}$ and $F_{rect}$: 
\begin{equation}
\begin{aligned}
    L_{\text{align}} &= \frac{1}{C} \sum_{c} \left( \left| \mu\left( F_{o} \right) - \mu\left( F_{\text{rect}} \right) \right| \right. \\
    &\quad + \left. \left| \sigma\left( F_{o} \right) - \sigma\left( F_{\text{rect}} \right) \right| \right).
\end{aligned}
\end{equation}

We optimize the model with the final loss $L$:
\begin{equation}
\begin{aligned}
    \mathrm {} L = L_{BCE} +L_{cyc}+L_{align}
\end{aligned}
\end{equation}

\section{Experiments}

\begin{table*}[!t]
    \centering
    \caption{ Mean-IoU of 1-way 1-shot and 5-shot results of tradictional few-shot approaches and cross-domain few-shot method on the four CD-FSS benchmark.Bold denotes the best performance
among all methods.}
    \vspace{-0.1in}
    \label{table:comparison}
    \scalebox{0.95}{
    \begin{tabular}{@{}ccccccccccc@{}}
\toprule
\multirow{2}{*}{Methods}                                                                                                  & \multicolumn{2}{c|}{Deepglobe}                                                                                                                                                                                                     & \multicolumn{2}{c|}{ISIC}                                                                                                                                                                                                          & \multicolumn{2}{c|}{Chest X-ray}                                                                                                                                                                                                   & \multicolumn{2}{c|}{FSS-1000}                                                                                                                                                                                                      & \multicolumn{2}{c}{Average}                                                                                                                                                                                   \\ \cmidrule(l){2-11} 
                                                                                                                          & \multicolumn{1}{c|}{1-shot}                                                                           & \multicolumn{1}{c|}{5-shot}                                                                                                & \multicolumn{1}{c|}{1-shot}                                                                           & \multicolumn{1}{c|}{5-shot}                                                                                                & \multicolumn{1}{c|}{1-shot}                                                                           & \multicolumn{1}{c|}{5-shot}                                                                                                & \multicolumn{1}{c|}{1-shot}                                                                           & \multicolumn{1}{c|}{5-shot}                                                                                                & \multicolumn{1}{c|}{1-shot}                                                                           & 5-shot                                                                                                \\ \midrule
\multicolumn{11}{c}{Few-shot Segmentation Methods}                                                                                                                                                                                                                                                                                                                                                                                                                                                                                                                                                                                                                                                                                                                                                                                                                                                                                                                                                                                                                                                                                                                                                                                                                                            \\ \midrule

\multicolumn{1}{l|}{\begin{tabular}[c]{@{}l@{}}PGNet~\cite{zhang2019pyramid}\\ PANet~\cite{wang2019panet}\\ CaNet~\cite{zhang2019canet}\\ RPMMs~\cite{yang2020prototype}\\ PFENet~\cite{tian2020prior}\\ RePRI~\cite{boudiaf2021few}\\ HSNet~\cite{min2021hypercorrelation}\\ SSP~\cite{fan2022self}\end{tabular}} & \begin{tabular}[c]{@{}c@{}}10.73\\ 36.55\\ 22.32\\ 12.99\\ 16.88\\ 25.03\\ 29.65\\ 40.48\end{tabular} & \multicolumn{1}{c|}{\begin{tabular}[c]{@{}c@{}}12.36\\ 45.43\\ 23.07\\ 13.47\\ 18.01\\ 27.41\\ 35.08\\ 49.66\end{tabular}} & \begin{tabular}[c]{@{}c@{}}21.86\\ 25.29\\ 25.16\\ 18.02\\ 23.50\\ 23.27\\ 31.20\\ 35.09\end{tabular} & \multicolumn{1}{c|}{\begin{tabular}[c]{@{}c@{}}21.25\\ 33.99\\ 28.22\\ 20.04\\ 23.83\\ 26.23\\ 35.10\\ 44.96\end{tabular}} & \begin{tabular}[c]{@{}c@{}}33.95\\ 57.75\\ 28.35\\ 30.11\\ 27.22\\ 65.08\\ 51.88\\ 74.23\end{tabular} & \multicolumn{1}{c|}{\begin{tabular}[c]{@{}c@{}}27.96\\ 69.31\\ 28.62\\ 30.82\\ 27.57\\ 65.48\\ 54.36\\ 80.51\end{tabular}} & \begin{tabular}[c]{@{}c@{}}62.42\\ 69.15\\ 70.67\\ 65.12\\ 70.87\\ 70.96\\ 77.53\\ 79.03\end{tabular} & \multicolumn{1}{c|}{\begin{tabular}[c]{@{}c@{}}62.74\\ 71.68\\ 72.03\\ 67.06\\ 70.52\\ 74.23\\ 80.99\\ 80.56\end{tabular}} & \begin{tabular}[c]{@{}c@{}}32.24\\ 47.19\\ 36.63\\ 31.56\\ 34.62\\ 46.09\\ 47.57\\ 57.20\end{tabular} & \begin{tabular}[c]{@{}c@{}}31.08\\ 55.10\\ 37.99\\ 32.85\\ 34.98\\ 48.34\\ 51.38\\ 63.92\end{tabular} \\ \midrule
\multicolumn{11}{c}{Cross-domain Few-shot Segmentation Methods}                                                                                                                                                                                                                                                                                                                                                                                                                                                                                                                                                                                                                                                                                                                                                                                                                                                                                                                                                                                                                                                                                                                                                                                                                               \\ \midrule
\multicolumn{1}{l|}{\begin{tabular}[c]{@{}l@{}}PATNet~\cite{lei2022cross}\\ Ours\end{tabular}}                                                & \begin{tabular}[c]{@{}c@{}}37.89\\ \textbf{41.29}\end{tabular}                                                 & \multicolumn{1}{c|}{\begin{tabular}[c]{@{}c@{}}42.97\\ \textbf{50.12}\end{tabular}}                                                 & \begin{tabular}[c]{@{}c@{}}\textbf{41.16}\\ 40.77\end{tabular}                                                 & \multicolumn{1}{c|}{\begin{tabular}[c]{@{}c@{}}\textbf{53.58}\\ 48.87\end{tabular}}                                                 & \begin{tabular}[c]{@{}c@{}}66.61\\ \textbf{82.35}\end{tabular}                                                 & \multicolumn{1}{c|}{\begin{tabular}[c]{@{}c@{}}70.20\\ \textbf{82.31}\end{tabular}}                                                 & \begin{tabular}[c]{@{}c@{}}78.59\\ \textbf{79.05}\end{tabular}                                                 & \multicolumn{1}{c|}{\begin{tabular}[c]{@{}c@{}}\textbf{81.23}\\ 80.40\end{tabular}}                                                 & \begin{tabular}[c]{@{}c@{}}56.06\\ \textbf{60.86}\end{tabular}                                                 & \begin{tabular}[c]{@{}c@{}}61.99\\ \textbf{65.42}\end{tabular}                                                 \\ \bottomrule
\end{tabular}
    }
    \vspace{-0.1in}
\end{table*}

\subsection{Datasets}
Following~\cite{lei2022cross}, we validate our methods on the cross-domain few-shot segmentation (CD-FSS) benchmark. This benchmark includes images and pixel-level annotations from the FSS-1000~\cite{li2020fss}, DeepGlobe~\cite{demir2018deepglobe}, ISIC2018~\cite{codella2019skin,tschandl2018ham10000}, and Chest X-ray datasets~\cite{candemir2013lung,jaeger2013automatic}. These datasets range from natural to medical images, providing sufficient domain diversity. We train models on the natural image dataset PASCAL VOC 2012~\cite{everingham2010pascal} with SBD~\cite{hariharan2011semantic} augmentation and evaluate models on the CD-FSS benchmark.

\textbf{FSS-1000}~\cite{li2020fss} is a dataset designed for few-shot segmentation, containing 1,000 different categories of natural objects and scenes, with each category comprising 10 annotated images. We evaluate models on the official test set with 2,400 images.

\textbf{Deepglobe}~\cite{demir2018deepglobe} is a complex Geographic Information System (GIS) dataset, containing satellite images with categories of urban, agriculture, rangeland, forest, water, barren, and unknown. For testing, we follow the processing in~\cite{lei2022cross} to divide each image into six patches and filtering out the 'unknown' category, and evaluate models on the resulting 5,666 test images and their corresponding masks.

\textbf{ISIC2018}~\cite{codella2019skin,tschandl2018ham10000} is used for skin lesion analysis, containing numerous skin images with associated segmentation labels. We evaluate models on the official training set following the common practice~\cite{codella2019skin}, using a uniform resolution of 512×512 pixels, comprising a total of 2,596 test images.

\textbf{Chest X-ray}~\cite{candemir2013lung,jaeger2013automatic} is an X-ray image dataset for tuberculosis detection, containing X-ray images of Tuberculosis cases as well as images from normal cases. We downsample the original image resolution to 1024×1024 pixels for testing.

\subsection{Implementation Details}
We utilize the train set of PASCAL VOC dataset as the source domain training set. During training, we employ SSP~\cite{fan2022self} with the ResNet-50 backbone as the baseline model. 
We first train the baseline model on the whole training set, and then train our method with additional 5 epochs using a batch size of 8. We use SGD to optimize our model, with a 0.9 momentum and an initial learning rate of 1e-3. To reduce memory consumption and accelerate the training process, we resize both query and support images to 400 × 400. We apply our two domain perturbation modules into the first three layers of ResNet. For local perturbations, we use the Gaussian noise with a mean of zero and a standard deviation of 0.75, while for global perturbations, we used the Gaussian noise with a mean of zero and a standard deviation of one. 
All models are evaluated using the mean Intersection Over Union (mIOU).

\subsection{Comparison Experiments}
In Table \ref{table:comparison}, we present a comparison between our method and other approaches, including traditional few-shot segmentation methods and existing cross-domain few-shot segmentation methods. Traditional few-shot segmentation methods usually underperform in cross-domain scenarios due to the large domain gap between the train and test data. While our approach effectively reduces the domain gap and improves the segmentation performance. This performance improvement is particularly notable in Chest X-rays, where our 1-shot and 5-shot performance surpasses the PATNet~\cite{lei2022cross} by 15.74\% and 12.11\%, respectively. In Deepglobe, the improvement is 3.4\%(1-shot) and 7.15\%(5-shot). For FSS-1000, our model achieves comparable performance to PATNet, because the domain gap is small. 

We also follow the same setting of RD \cite{wang2022remember} to train our model on VOC and evaluate models on SUIM.
Table \ref{table:RD} shows our method performs much better than RD.

\begin{table}[!t]
\centering
\caption{Mean-IoU of 1-way 1-shot results of our method following the same setting of RD\cite{wang2022remember}.}
    \vspace{-0.1in}
\label{table:RD}
\small
\resizebox{0.475\textwidth}{!}{%
\begin{tabular}{@{}l|c|c|c|c|c@{}}
\toprule
\multicolumn{1}{c|}{} & split-0 & split-1 & split-2 & split-3 & Average  \\ \midrule
RD\cite{wang2022remember} & 35.20    & 33.40    & 34.30    & 36.00      & 34.70\\
Ours & 40.60   & 38.18   & 41.53   & 40.72   & 40.25 \\ \bottomrule
\end{tabular}
    \vspace{-0.15in}
}
\end{table}

\begin{figure}[t]
  \centering
   \includegraphics[width=\linewidth]{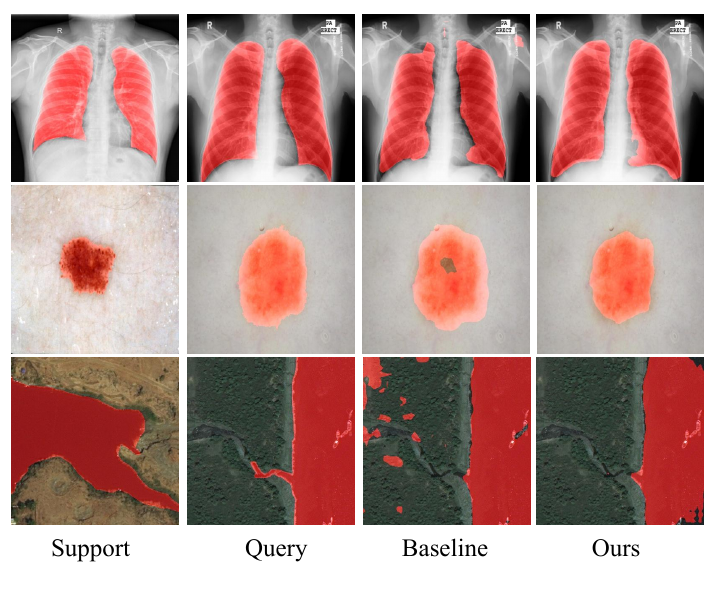}
   \vspace{-0.3in}
   \caption{Qualitative results of our model and baseline in 1-way 1-shot setting on challenging scenarios with large domain gap.}
   \label{fig:qualitative}
   \vspace{-0.15in}
\end{figure}

We present some qualitative results of our proposed model for 1-way 1-shot segmentation in Fig. \ref{fig:qualitative}. These results indicate that our method improves the generalization ability of traditional few-shot models, attributing to its capability of aligning various domains to the source domain.

\begin{figure}[!t]
  \centering
   \includegraphics[width=0.85\linewidth]{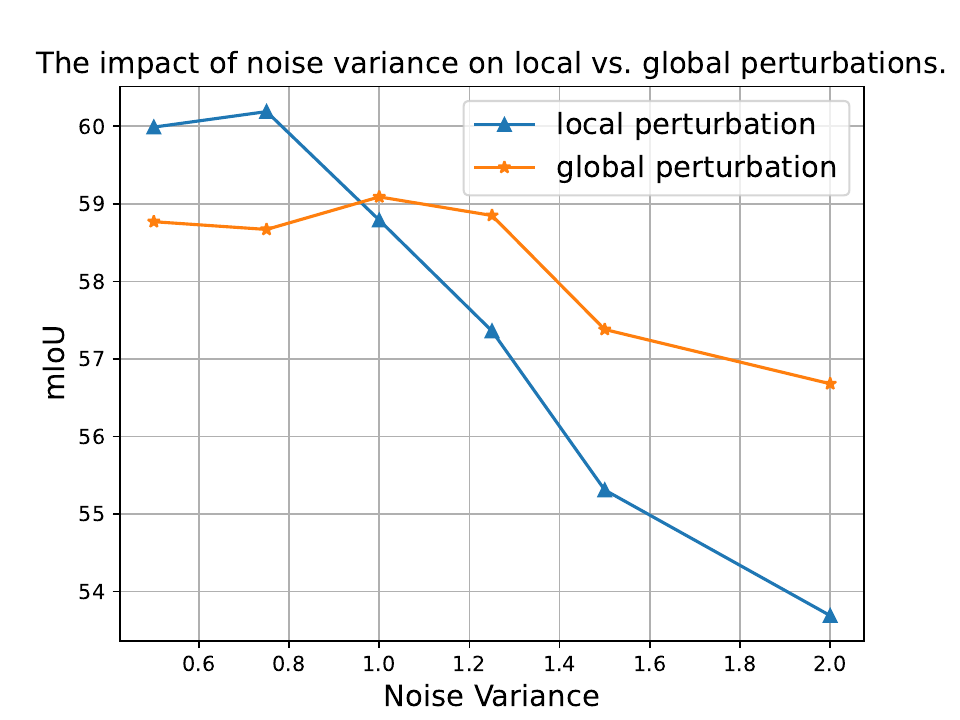}
    \vspace{-0.1in}
   \caption{We demonstrate the trend of global and local perturbations under different Gaussian noise variances.}
   \label{fig:noise_impact}

\end{figure}

\begin{table}[]
\small
\caption{The effects of each module within the baseline, namely the Perturbation module, Rectification module, and Cyclic Alignment Loss, are demonstrated.}
    \vspace{-0.1in}
\label{table:each_module}
\resizebox{0.475\textwidth}{!}{%
\begin{tabular}{@{}ccc|c@{}}
\toprule
Perturbation & Rectification & Cyclic Alignment & mean-IoU \\ \midrule
             &               &                  & 57.20    \\
\checkmark            &               &                  & 58.45    \\
             & \checkmark             &                  & 57.80    \\
\checkmark            & \checkmark             &                  & 59.17    \\
\checkmark            & \checkmark             & \checkmark                & 60.86    \\ \bottomrule
\end{tabular}
}

\end{table}

\subsection{More Analysis}
We conduct extensive ablation experiments to demonstrate and analyze the effectiveness of our approach. 

\subsubsection{Ablation Studies}
We conduct comprehensive ablation experiments to evaluate the effectiveness of our proposed components.

\begin{table}[]
\small
\caption{Results of using our Cyclic Alignment Loss.}
    \vspace{-0.1in}
\label{table:CLloss}
\resizebox{0.475\textwidth}{!}{
\begin{tabular}{cccc}
  \toprule
  BCE loss & + cyclic loss &  + align loss  & + cyclic \& align loss \\
  \midrule
  57.65 & 58.56 & 59.12 & 60.86 \\
  \bottomrule

  \end{tabular}
}
    \vspace{-0.05in}

\end{table}

\textbf{Impact of Noise Variance on Perturbations.} 
Figure \ref{fig:noise_impact} illustrates the effects of Gaussian noise with varying variances in local and global perturbations. Local perturbations suffer from performance degradation with slightly higher noise levels, whereas global perturbations withstand larger noise levels with minimal performance impact, suggesting greater stability. Thus, in our method, we set Gaussian noise variances at 0.75 for local and 1 for global perturbations to broaden the simulated domain range and improve the domain generalizability.

\textbf{Impact of Each Component.} 
Table~\ref{table:each_module} illustrates the effectiveness of each module in the model. Integration of all modules results in 3.66\% performance improvement compared to the SSP~\cite{fan2022self} baseline. Importantly, the feature perturbation and rectification processes complement each other: perturbation simulates features across domains, and rectification aligns these features back to the source domain space. Solely using feature perturbation degrades the model to the domain generalization approach similar to NP~\cite{fan2022towards}. 
Additionally, our cyclic alignment loss is indispensable as it ensures the unification of images from various domains to the source domain.

Table \ref{table:CLloss} shows the ablation analysis on Cyclic Alignment Loss.
Both the alignment loss and cyclic loss can improve the performance.

\begin{figure}[!t]
  \centering
   \includegraphics[width=0.9\linewidth]{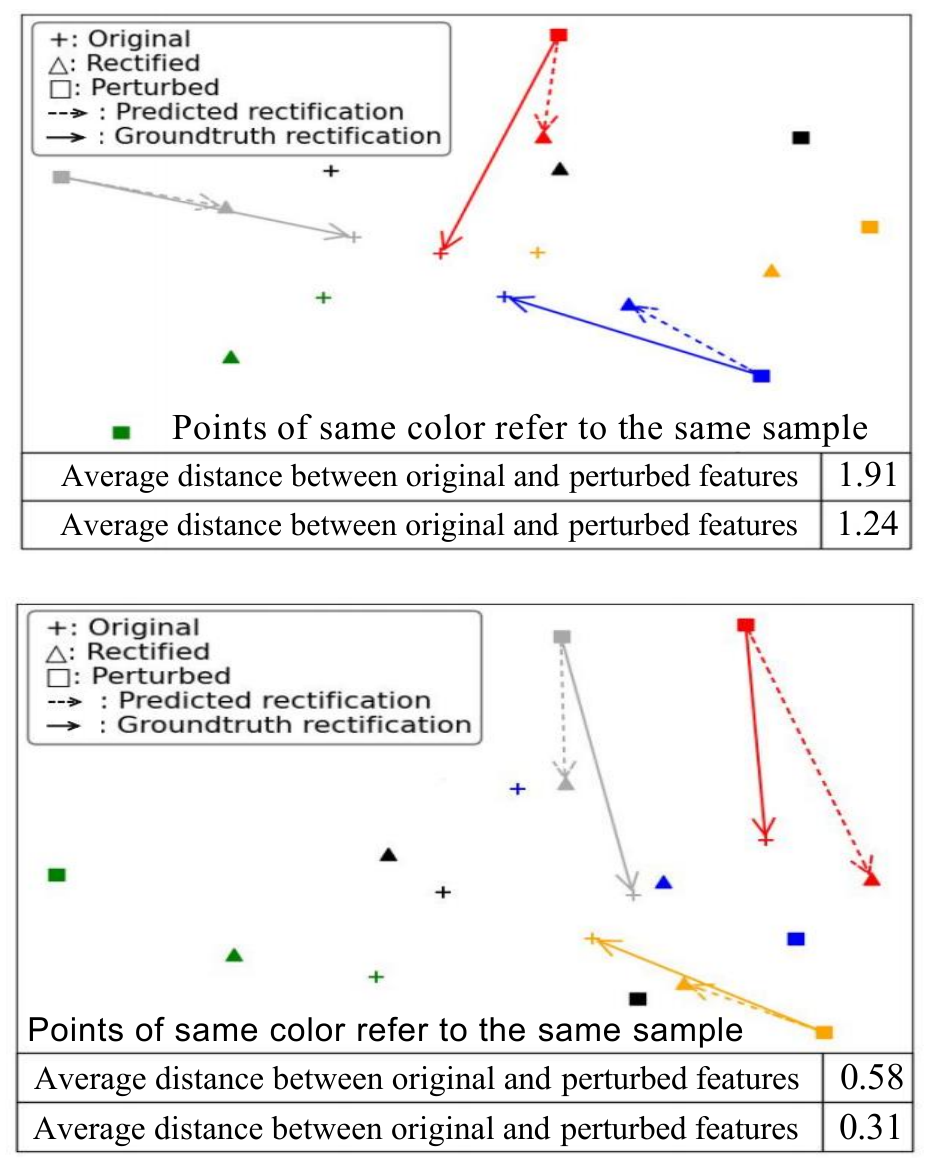}
    \vspace{-0.1in}
   \caption{Visual analysis (t-SNE) of channel-wise means(top) and variations(bottom).}
   \label{fig:tsne}

\end{figure}

Table~\ref{table:Perturbation_way} compares the impact of local and global styles in feature perturbation, showing that their combination improves model performance attributing to a wider range of domain simulation.
Using the channel-wise means and variances as features, the t-SNE(Figure~\ref{fig:tsne}) shows that the perturbed features are rectified to be closer to the original features, demonstrating our model's effectiveness.

\begin{table}[]
\small
\caption{Results of feature perturbation methods.}
    \vspace{-0.1in}
\label{table:Perturbation_way}
\resizebox{0.475\textwidth}{!}{%
\begin{tabular}{@{}c|ccc@{}}
\toprule
         & local style & global style & Both perturbation \\ \midrule
mean-IoU & 59.81        & 59.17         & 60.86       \\ \bottomrule
\end{tabular}
}
    \vspace{-0.05in}

\end{table}

\textbf{Impact of Noise Types.} 
We choose the popular Gaussian distribution to generate random noises, which has been widely used by other works (\eg, Mixstyle, DSU and NP).
Perturbing feature statistics with random noises can effectively synthesize diverse domain styles, while the noise type is not essential.
Table~\ref{table:Noise_type} shows that our method is insensitive to the noise types, performing well with Beta, and Uniform noises.
Note that our novel Local-Global Domain Perturbation and Cyclic Domain Alignment can largely improve the domain style synthesis diversity for all kinds of noise. 

\begin{table}[]
\small
\caption{Results of using different types of noise.}
    \vspace{-0.1in}
\label{table:Noise_type}
\resizebox{0.475\textwidth}{!}{
\begin{tabular}{@{}c|ccc@{}}
\toprule
Noise Type & Gaussian (1,0.75) & Beta (3,4)   & Uniform (-1,1) \\ \midrule
mIoU & 60.86      & 60.78 & 60.00       \\ \bottomrule
\end{tabular}
}

\end{table}

\textbf{More adapters.} 
Table~\ref{table:rounds} shows that applying multiple adapters can further improve performance.

\begin{table}[!t]
\centering
\caption{Results of using one/two adapters within a single stage.}
    \vspace{-0.1in}
\label{table:rounds}
\small
\resizebox{0.475\textwidth}{!}{
\begin{tabular}{@{}l|c|c|c|c|c@{}}
\toprule
\multicolumn{1}{c|}{} & FSS   & Chest & Deepglobe & ISIC & Average  \\ \midrule
One adapter & 79.05 & 82.35 & 41.29     & 40.77 & 60.86\\
Two adapters & 79.25 & 83.04 & 41.74 & 41.63 & 61.41 \\ \bottomrule
\end{tabular}
}

\end{table}


\begin{table}[t]
\small
\centering
\caption{ Comparison to domain adaption and domain generalization approaches under 1-shot setting. We use same baseline with different methods to ensure fair comparison.}
    \label{table:with_DA_DG}
    \vspace{-0.1in}
\resizebox{0.475\textwidth}{!}{%
\begin{tabular}{@{}l|c|c|c|c|c@{}}
\toprule
Method   & FSS   & Chest & Deepglobe & ISIC  & Average \\ \midrule
Baseline(SSP~\cite{fan2022self}) & 79.03 & 74.23 & 40.48     & 35.09 & 57.20   \\
AdaIN~\cite{huang2017arbitrary}    & 78.89 & 74.23 & \textbf{41.85}     & 34.36 & 57.33   \\
Mixstyle~\cite{zhou2021domain} & \textbf{79.24} & 76.63 & 41.05     & 35.98 & 58.21   \\
DSU~\cite{li2022uncertainty}      & 78.99 & 77.83 & 41.19     & 36.64 & 58.66   \\
NP~\cite{fan2022towards}       & 78.98 & 76.44 & 41.83     & 37.87 & 58.78   \\
Ours     & 79.05 & \textbf{82.35} & 41.29     & \textbf{40.77} & \textbf{60.86}   \\ \bottomrule
\end{tabular}
}

\end{table}

\subsubsection{Comparion with Domain Transfer Methods}
We compare our method against traditional domain adaptation (DA) and domain generalization (DG) approaches to validate our method's effectiveness. For a fair comparison, all categories in the PASCAL VOC were used for training in both DA and DG methods. 
We evaluate models in the 1-shot setting on the CD-FSS benchmark.

\textbf{Domain Adaptation.} We adopt the classical AdaIN~\cite{huang2017arbitrary} method to train four models for the four test datasets. During training, we randomly sample images from the test dataset and extract their feature channel statistics in the low-level feature map. And then the AdaIN is applied to replace the feature channel statistics of the train image with the extracted statistics from the test dataset.

\textbf{Domain Generalization.} We employ the Mixstyle~\cite{zhou2021domain}, DSU~\cite{li2022uncertainty} and NP~\cite{fan2022towards} methods for comparison. These approaches also involves perturbing feature statistics, but they only perform local perturbations and lack a feature rectification process. 

Table \ref{table:with_DA_DG} shows that our method performs much better than DA and DG methods in cross-domain few-shot segmentation.

\subsubsection{Applying SAM in CD-FSS}

The recent released large-scale SAM~\cite{kirillov2023segment} model has greatly advanced image segmentation, demonstrating remarkable zero-shot segmentation capabilities.
However, SAM cannot be directly applied to cross-domain few-shot segmentation.
Thus we evaluate PerSAM~\cite{zhang2023personalize} to compare our method to the SAM-based method in cross-domain few-shot segmentation.
PerSAM is a training-free method. It adapts SAM into the one-shot setting by using support images as the prompt input to segment target objects in query images.
Table~\ref{table:PerSAM} shows that our method performs much better than PerSAM in cross-domain few-shot segmentation. 

\subsubsection{Extension to Transformer}
In Table \ref{table:with_transformer}, we show the results of applying our method within FPTrans\cite{zhang2022feature}, which leverages support sample prototypes as prompts and Vision Transformer (ViT) as the backbone. Applying our method to the lower-level blocks of ViT improves performance in cross-domain datasets. 

\begin{table}[!t]
\centering
\caption{The result of directly applying PerSAM to cross-domain few-shot segmentation.}
    \label{table:PerSAM}
    \vspace{-0.1in}
\small
\resizebox{0.475\textwidth}{!}{
\begin{tabular}{@{}l|c|c|c|c|c@{}}
\toprule
\multicolumn{1}{c|}{} & FSS   & Chest & Deepglobe & ISIC & Average  \\ \midrule
PerSAM~\cite{zhang2023personalize}                & 79.65 & 31.12 & 33.39     & 21.27 & 41.35\\
Ours~                & 79.05 & 82.35 & 41.29     & 40.77 & 60.86 \\ \bottomrule
\end{tabular}
}
\end{table}

\begin{table}[]
\small
\caption{Applying our method to transformers can further enhance the model's performance in cross-domain tasks.}
    \vspace{-0.1in}
\label{table:with_transformer}
\resizebox{0.475\textwidth}{!}{%
\begin{tabular}{@{}l|c|c|c|c|c@{}}
\toprule
\multicolumn{1}{c|}{} & FSS   & Chest & Deepglobe & ISIC & Average  \\ \midrule
FPTrans~\cite{zhang2022feature}                & 78.92 & 80.49 & 39.21     & 47.79 & 61.60\\
FPTrans + ours~                & 78.63 & 82.74 & 40.32     & 49.43 & 62.78 \\ \bottomrule
\end{tabular}

}
\end{table}

\section{Conclusion}

In this paper, we propose a method to effectively bridge the domain gap between different datasets by aligning the target domain with the source domain space.
During training, we train a unified adapter by using simulated perturbed features. In the inference stage, we consider target domain images as a form of perturbed images for the direct rectification.
Furthermore,we introduce both local and global perturbations to ensure significant style changes, not only based on individual sample but also on the overall style of the dataset.
We utilize a cyclic alignment loss to ensure the alignment between the source and target domains for model optimization.
We conduct extensive experiments to validate the effectiveness of the proposed framework on various cross-domain segmentation tasks and achieve state-of-the-art (SOTA) results on multiple benchmarks.

\textbf{Acknowledgement.} This work was supported in part by the National Natural Science Foundation of China (U2013210, 62372133),  Guangdong Basic and Applied Basic Research Foundation under Grant (Grant No. 2022A1515010306, 2024A1515011706), Shenzhen Fundamental Research Program (Grant No. JCYJ20220818102415032), and Shenzhen Key Technical Project (Grant NO. KJZD20230923115117033).

{
    \small
    \bibliographystyle{ieeenat_fullname}
    \bibliography{main}
}

\end{document}